\documentclass[onecolumn]{IEEEtran}
\IEEEoverridecommandlockouts

\usepackage[ruled,vlined]{algorithm2e}

\usepackage[utf8]{inputenc}
\usepackage[T1]{fontenc}

\usepackage{amsmath}
\usepackage{amsfonts}       
\usepackage{amssymb}
\usepackage{hyperref}

\usepackage{graphicx}       
\usepackage{subcaption}  
\usepackage{float}

\usepackage{array}          
\usepackage{tabularx}
\usepackage{multirow}

\usepackage{xcolor}
\usepackage{tikz}
\usepackage{pifont} 
\usepackage{tikz}
\usetikzlibrary{automata,positioning}
\usepackage[
top    = 0.75in,
bottom = 1.2in,
left   = 0.75in,
right  = 0.75in]
{geometry}

\newcommand{\cmark}{\textcolor{green}{\ding{51}}} 
\newcommand{\xmark}{\textcolor{red}{\ding{55}}}   
\definecolor{sdr}{rgb}{0.0, 0.65, 0.31}

\def\BibTeX{{\rm B\kern-.05em{\sc i\kern-.025em b}\kern-.08em
    T\kern-.1667em\lower.7ex\hbox{E}\kern-.125emX}}

\renewcommand{\baselinestretch}{.90}

\begin{document}

\title{Reliable Multi-view 3D Reconstruction for `Just-in-time' Edge Environments\thanks{This material is based upon work supported by the National Science Foundation (NSF) under Award Numbers: CNS-1943338 and CNS-2401928}}


\author{
\IEEEauthorblockN{
Md. Nurul Absur\IEEEauthorrefmark{1},
Abhinav Kumar\IEEEauthorrefmark{2}
Swastik Brahma\IEEEauthorrefmark{2}
Saptarshi Debroy\IEEEauthorrefmark{1}}
\IEEEauthorblockA{
\IEEEauthorrefmark{1}City University of New York, USA; 
\IEEEauthorrefmark{2}University of Cincinnati, USA\\
Emails: 
\IEEEauthorrefmark{1}mbsur@gradcenter.cuny.edu;
\IEEEauthorrefmark{2}\{kumar3a3,brahmask\}@ucmail.uc.edu; 
\IEEEauthorrefmark{1}saptarshi.debroy@hunter.cuny.edu
}
}

\maketitle

\begin{abstract}
Multi-view 3D reconstruction 
applications are revolutionizing critical use cases that require rapid situational-awareness, such as emergency response, tactical scenarios, and public safety. In many cases, their near-real-time latency requirements and ad-hoc needs for compute resources necessitate adoption of `Just-in-time' edge environments where the system is set up on the fly to support the applications during the mission lifetime. 
However, reliability issues can arise from the inherent dynamism and operational adversities of such edge environments, resulting in spatiotemporally correlated disruptions that impact the camera operations, which can lead to sustained degradation of reconstruction quality.
In this paper, we propose a novel portfolio theory inspired edge resource management strategy for reliable multi-view 3D reconstruction against possible system disruptions. 
Our proposed methodology can guarantee reconstruction quality satisfaction even when the cameras are prone to spatiotemporally correlated disruptions. The portfolio theoretic optimization problem is solved using a genetic algorithm that converges quickly for realistic system settings. 
Using publicly available and customized 3D datasets, we demonstrate the proposed camera selection strategy's benefits in guaranteeing reliable 3D reconstruction against traditional baseline strategies, under spatiotemporal disruptions.
\end{abstract}


\section{Introduction}

Multi-view 3D reconstruction~\cite{10419312} applications are increasingly becoming fundamental for complex 3D video processing application pipelines that create immersive 3D environments, such as, virtual, augmented, and mixed reality~\cite{8998140}. For multi-view 3D reconstruction, 2D images of a scene are simultaneously captured by cameras from multiple viewpoints and then fused together to get a 3D visualization. In many cases, such applications are hosted by critical use cases, such as emergency response, tactical scenarios, and public safety, that rely on rapid and high quality reconstruction results for mission success. In order to provide such rapid situational awareness, these use cases often rely on `just-in-time' (JIT) edge environments where: i) raw video data is captured by end-devices, such as drones and robots; ii) specialty on-premise 
edge servers equipped with CPU/GPU and 3D video processing capabilities (which are sometimes hosted on multi-utility vehicles), process the raw video data to produce reconstructed scenes in 3D;
and iii) ground consumers (e.g., tactical or first responder units) visualize the reconstructed scene on their hand-held devices.

The already tricky proposition of resource management in JIT edge environments becomes even more challenging as such ad-hoc system deployments with weakly-coupled components are often prone to `disruptions' that can have potentially catastrophic impact the reconstruction quality. In general, such disruptions may occur due to various unintended factors, such as, abrupt changes in the operating environment~\cite{8922760}, and/or malicious factors, such as jamming attacks on wireless channels~\cite{7009348}.  Although such disruptions can affect any edge system component (e.g., servers, network, cameras), their impact (i.e., in terms of reconstructed outcome) are most profound on the cameras and/or camera-enabled devices due to lack of redundancy (in terms of their numbers) within JIT edge environments and the relative importance of each and every camera (and its viewpoint) in generating high quality reconstruction.

These disruptions, triggered by the aforementioned causes can lead to: i) `spatially correlated' failures in many cameras, impacting multiple or all at the same time and/or ii) a series of `temporally correlated' failures across many cameras over a period of time. An example of spatially correlated disruption can be a sudden obstruction of view of multiple cameras due to smoke in an indoor fire emergency event, leading to lack of useful data (from such cameras) towards 3D reconstruction.
Such lack of useful data even from a single camera can often result in significant reconstruction quality degradation, leading to useless reconstruction. 
{\em Thus, there is a need for reliable camera management in the presence of such spatiotemporally correlated disruptions to ensure minimum reconstruction quality satisfaction. Although there exist strategies for mission critical edge resource management~\cite{10.1145/3589639}, little work is done to ensure reliability in such environments in the presence of disruptions.} 



In this paper, we propose an intelligent JIT edge resource management technique that makes multi-view 3D reconstruction in such environments reliable under system-wide disruptions. 
In particular, our methodology involves an innovative camera selection strategy inspired by 
{\em portfolio theory} that is aimed at mitigating the potential catastrophic effects of spatial and temporal disruptions that are commonplace in JIT edge environments~\cite{owhadikareshk2021portfolio}. First through benchmarking experiments with 3D reconstruction datasets, we demonstrate the importance and non-triviality of camera selection towards useful 3D reconstruction under disruptions. Then we formulate the camera selection problem for minimum reconstruction quality satisfaction as a portfolio theory-inspired investment fluctuation minimization problem. 
Next, we propose a genetic algorithm-based solution to the optimization problem that demonstrates rapid convergence in realistic scenarios, ensuring practical utility. 


We evaluate the proposed edge resource management technique through a detailed simulation that mimics spatiotemporally correlated disruptions on JIT edge environments. Using both our own and publicly available multi-view 3D reconstruction datasets and state-of-the-art 3D reconstruction pipeline, we run extensive experiments to evaluate and compare the quality of 3D reconstruction in 3 ways: i) without disruptions, ii) under disruptions but addressed using traditional baseline approaches towards ensuring reliability, and iii) under disruption but resolved using our proposed portfolio theoretic approach. The results demonstrate that under different degrees of system-wide disruptions and camera selection stipulations, our proposed method can ensure more reliable reconstruction in terms of higher mean value and lower standard deviation of reconstruction quality than baseline strategies. The results also overwhelmingly manifest our proposed approach's superiority in guaranteeing useful reconstructed results than traditional baseline approaches, under disruptions. Overall, the results validate our theoretical constructs and emphasize the practical enhancements that our proposed approach offers to mission-critical 3D reconstruction applications hosted by JIT edge environments.      



The rest of the paper is organized as follows. Section~\ref{sec:related-work} discusses the related work in this area. Section~\ref{sec:background} introduce background information on multi-view 3D reconstruction and experimentally analyses problem evidence. Section~\ref{sec:sys-model} describes the system model and formulates the portfolio theoretic optimization problem. 
Section~\ref{sec:finalevaluation} discusses performance evaluation. Section~\ref{sec:conclusions} concludes the paper.

\section{Related Work}
\label{sec:related-work}

Disruption management in distributed and networked systems to improve system reliability and security has been an active area of research for sometime. Jeong et al. \cite{8685918} discussed the importance of reducing service disruption by introducing the ReSeT system, which reduces the handover time of mobile nodes. 
In \cite{9339378}, the authors reflect on major critical analyses of edge computing regarding quality, reliability, and latency. 
Li et al. \cite{8643392} reflect on the importance of latency management for real-life applications and propose a two-phase scheduling strategy to mitigate disruptions in edge environments.
An auction theory inspired strategy is proposed in \cite{8812207} to tackle disruptions caused by improper resource management and latency constraints. 

There have some significant contributions in edge resource management for applications' latency satisfaction, however very few with the focus of disruption management. 
Among the notable works in this space, in \cite{cheng2023resilient}, the authors propose a novel two-stage adaptive robust system to ensure resilience against any unpredictable failures. 
Aral et al. \cite{8567669} explore possible failures 
while achieving quality of service (QoS) properties, such as geographical prevalence and low latency. 
In \cite{8780405}, the authors explore all  possible threats that can degrade system efficiency, and propose a resilient security solution for IoT systems.
The authors in \cite{10138654} propose an adaptive distributed inference algorithm that can outperform traditional algorithms in terms of resilience against failures.


Although rare, in recent times, portfolio theory has been  applied to computing science and engineering problems.
In \cite{owhadikareshk2021portfolio}, the authors propose PortFawn, an open-source Python library that is capable of managing summery-variance portfolio optimization approaches to reduce the investment risk of the assets. 
There are some work that combine portfolio and optimization theories for computing problems. 
In \cite{7877194}, a combination of three genetic algorithm variants is proposed to show better performance while solving portfolio problems with constraints.
The authors in \cite{SHAVERDI2020105892} demonstrate the benefits of portfolio theory by designing a multi-objective and robust probabilistic model for portfolio optimization. 
Another significant application of portfolio theory in computing is \cite{9148668} where portfolio theory is used for traffic allocation in 5G heterogeneous cloud-RAN to achieve better quality of experience. 
{\em In most of the aforementioned work, portfolio theory is applied to optimization problems where the underlying application is mostly finance and economics. However, there is a lack of literature where the principles of balanced portfolio management are applied to explore the possibilities of reliability management in cyber infrastructures, which this work seeks to propose.}


\section{Problem Evidence Analysis}
\label{sec:background}

\begin{figure*}[t]
    \centering
    \includegraphics[width=0.85\textwidth]{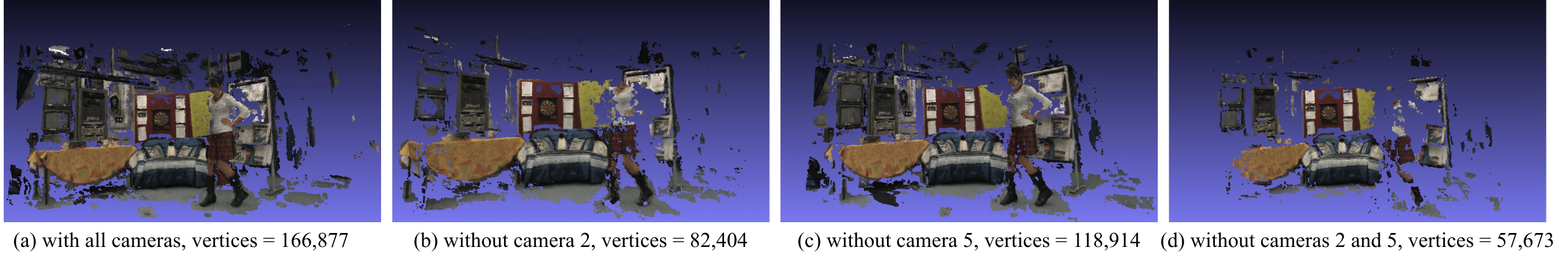}
    \caption{Effect of camera selection on multi-view 3D reconstruction quality}     
    \label{Fig:evidence}     
    \vspace*{-18pt}
\end{figure*}

Traditional multi-view 3D reconstruction techniques for dynamic scenes, such  as SfM+MVS pipelines~\cite{moulon2016openmvg,9904285}, use 
multiple cameras 
to estimate the 3D information of the captured scene. Given a set of high quality (i.e., high resolution) images observing a scene from different viewpoints with mutually overlapping regions, feature points, defined by a feature descriptor data structure such as SIFT~\cite{Lowe2004DistinctiveIF}, are detected for each image and matched across images to signify the same 3D points.
Thus, with more cameras covering the 3D scene from different viewpoints, the chances of matching more of such 3D points increases, which in turn increases reconstruction quality.

The objective of problem evidence analysis is to 
inquire whether choosing different sets of cameras for reconstruction would yield reconstruction of different quality.
To achieve such, we use the most widely used SfM+MVS pipeline based algorithm, viz., openMVG/openMVS~\cite{moulon2016openmvg,9904285} and publicly available Dance1~\cite{Mustafa_2015_ICCV} dataset (7 synchronous video sequences from static cameras). 
It goes without saying that the characteristics of the following results are not predicated on the choice of reconstruction algorithm; reconstruction methods other than SfM+MVS pipelines, such as deep neural network (DNN) based algorithms~\cite{yao2018mvsnet}, tend to show similar characteristics of reconstruction quality versus selected cameras for reconstruction.


Fig.~\ref{Fig:evidence} demonstrate the qualitative results from visual perspective, while the quantitative aspect is underpinned by the number of 3D point cloud vertices generated after the reconstruction process under different camera configurations. 
Fig.~\ref{Fig:evidence}(a) showcases the optimal scenario where all cameras contribute to the reconstruction (i.e., scenarios where no disruption impacts the cameras), amassing 166,877 vertices, indicative of the most detailed and comprehensive model.
However, when under disruption and Camera\#2 is omitted, i.e., is unable send useful data for reconstruction, as depicted in Fig.~\ref{Fig:evidence}(b), there is a substantial reconstruction quality reduction in terms of both visual effect and the vertices count of 82,404. This illustrates Camera\#2's significant role in the 3D scene's coverage. 
The removal of Camera\#5, however, as shown in Fig.~\ref{Fig:evidence}(c), results in a vertices tally of 118,914, demonstrating a notable but less dramatic decline in visual details. Thus, one can argue that the role of Camera\#2 is more significant than Camera\#5 for generating high quality reconstruction with Dance1~\cite{Mustafa_2015_ICCV} dataset.
The combined exclusion of Camera\#2 and Camera\#5 in Fig.~\ref{Fig:evidence}(d) yields the lowest vertices count of 57,673, further underscoring the criticality of these cameras in the 3D reconstruction process. 
{\em These results highlight the necessity for reliable methods to assess the trade-offs between the number of cameras and the reconstruction quality. 
Thus, one can argue that reconstruction dependability in situations where the integrity of camera operation as part of a JIT edge environment might be compromised, like disaster areas or tactical scenarios, depends on the careful selection of cameras.
Creating methods for the best camera choice is a fascinating intellectual pursuit and a practical necessity because it directly impacts the ability to generate dependable, high-fidelity 3D models in challenging real-world situations. 
}

\section{System Model and Problem Formulation}
\label{sec:sys-model}

\begin{figure}[t]
  \centering
  \includegraphics[width=0.40\textwidth]{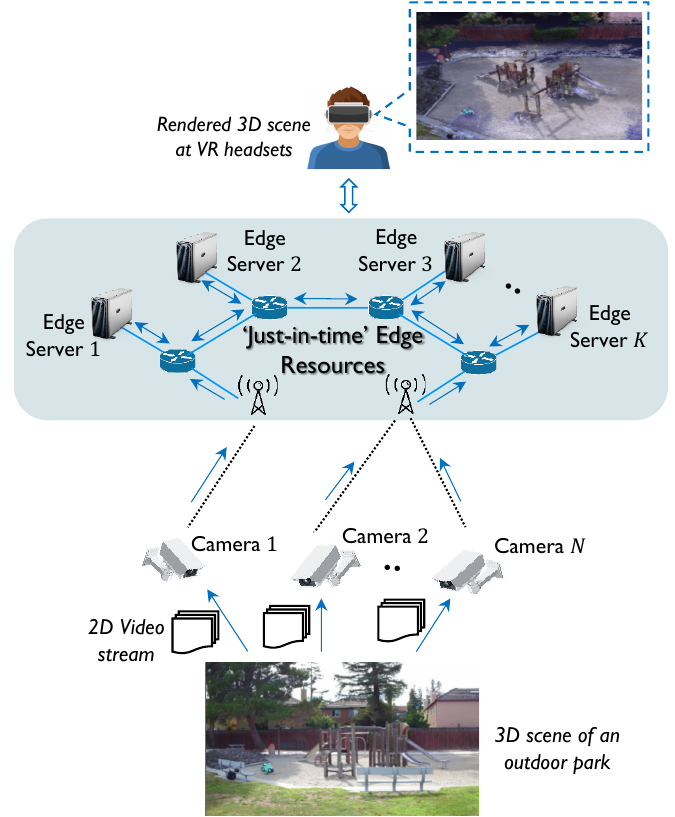}
  \caption{Multi-view 3D reconstruction of an outdoor scene hosted at JIT edge environment}
  \label{Fig:sys_model}
  \vspace{-0.2in}
\end{figure}

Our JIT edge system model, as depicted in Fig.~\ref{Fig:sys_model}, consists of a collection of $N$ camera-enabled devices, labeled as $\{1, \cdots, N\}$. In real-world implementations, such devices can be mobile. However, for this work, we assume static camera setting for simplicity of analysis. Each of these cameras capture an indoor or outdoor 3D scene (an outdoor park in our case) from different viewpoints in the form of a continuous stream of video frames. The video stream are transmitted to one or multiple edge servers ($\{1, \cdots, K\}$) in real-time where the video frames from different cameras captured at the same time-stamp are fused together using any state-of-the-art 3D reconstruction algorithm, to generate a 3D visualisation of the scene for that particular time-stamp. This process is repeated for all time-stamps and for the entire duration of the underlying application. 

For most JIT edge implementations, cameras and their supporting devices can be heterogeneous in nature, in terms of the device hardware, camera specifications, image resolutions, among other properties. This makes the problem of device/camera selection for reasonable reconstruction quality satisfaction under disruption, a non-trivial undertaking. 
In this work, we assume heterogeneity only in terms of camera image resolutions for simplicity of analysis, i.e., we consider that each camera $i$ is capable of sending video frames/images to edge servers at resolution $R_i$.  
Further, defining $p_{c_i}$ to be the probability with which camera $i$ is \textit{not} disrupted, we consider that a camera $i$, transmits images with specified image resolution $R_i$ with a probability $p_{c_i}$, and gets disrupted, producing no usable data, with the probability $1-p_{c_i}$.


To model \textit{fluctuations} of the disruption probabilities of the cameras, we consider that $p_{c_i}$ is a \textit{random variable}. Further, to model possible correlations between the disruptions experienced by
cameras $i$ and $j$, we consider that the correlation coefficient between $p_{c_i}$ and $p_{c_j}$ is $\rho-{i,j}$.
This correlation significantly influences the likelihood of simultaneous failures, thereby impacting the system's resilience.

\subsection{Portfolio Theoretic Formulation}
In such a scenario, it can be noted that the resolution of the image obtained from camera $i$, say denoted as $\mathcal{R}_i$, is itself a random variable, with $\mathcal{R}_i=0$ (when camera $i$ gets disrupted, which occurs with probability $p_{c_i}$)
and $\mathcal{R}_i=R_i$ (when camera $i$ is not disrupted), with $\mathbb{E}[\mathcal{R}_i] = R_i\mathbb{E}[p_{c_i}]$ (where $\mathbb{E}[\cdot]$ denotes expectation), which occurs with probability $p_{c_i}$).
Further, for cameras $i$ and $j$, with $p_{c_i}$ and $p_{c_j}$ being correlated random variables, with $\rho_{i,j}$ being their correlation coefficient, 
$\mathcal{R}_i$ and $\mathcal{R}_j$ also become correlated in nature 
whose covariance is $\mathrm{cov}(\mathcal{R}_i,\mathcal{R}_j) = R_i R_j \sigma_{p_{c_i}} \sigma_{p_{c_j}} \rho_{i,j}$, where $\sigma_{p_{c_i}}$ is the standard deviation of $p_{c_i}$.
Thus, the more correlated the disruptions of two cameras, the higher the covariance among the resolutions of the images 
obtained from them, 
and subsequently the higher the chances of the two cameras \textit{simultaneously} experiencing disruptions (and providing $0$ resolution images \textit{together}), which negatively
impacts performance.
To 
perform camera selection in such a scenario, suppose that 
camera $i$ is selected with a probability $\alpha_i$, with $\boldsymbol{\alpha} = (\alpha_1,\cdots,\alpha_N)$ denoting the camera selection vector, leading to the \textit{expected} total resolution of the images obtained from the cameras to be $\sum_{i=1}^N \alpha_i \mathbb{E}[\mathcal{R}_i]$. For simplicity of analysis, in this work, we consider $\alpha_i$ to be a binary variable where $\alpha_i = 1$ signifies selection of camera $i$ for reconstruction.

Portfolio theoretic resource management in the above problem scenario would enable selection of cameras such that a certain expected image quality (resolution) obtained from the cameras can be ensured (to sustain a desired quality of the reconstructed scene) while minimizing the covariances among the selected cameras’ image qualities to ensure reliability. Such a strategy would permit us to achieve optimal tradeoffs between the quality and reliability of a 3D reconstruction task in order to sustain its desired performance using disruption-prone limited system resources.
Specifically, the portfolio theoretic optimization problem for selecting cameras can be formulated
as~\eqref{cameraSelectOptProblem}.
\vspace{-0.1in}
\begin{subequations}
\footnotesize
\label{cameraSelectOptProblem}
\begin{align}
\underset{\boldsymbol{\alpha}=(\alpha_1,\ldots,\alpha_N)}{\text{minimize}} \quad &
\sum_{i = 1}^N \sum_{j=1}^N \alpha_i \alpha_j \,\mathrm{cov}(\mathcal{R}_i, \mathcal{R}_j) \tag{1a} \label{eq:PortfolioThPrelimObj} \\
\text{subject to} \quad &
\sum_{i=1}^N \alpha_i \,\mathbb{E}[\mathcal{R}_i] \geq \Theta
\quad\text{(\textit{quality constraint})} \tag{1b} \label{eq:qualityConstraint} \\
&
\sum_{i=1}^N \alpha_i \leq \Psi
\quad\text{(\textit{resource constraint})} \tag{1c} \label{eq:resourceConstraint} \\
&
\alpha_i \in [0,1],~~\forall i
\quad\text{(\textit{selection constraint})} \tag{1d} \label{eq:probabilityFeasibility}
\end{align}
\end{subequations}
\vspace{-0.2in}

Note that~\eqref{cameraSelectOptProblem} seeks to determine
the camera selection strategy $\boldsymbol{\alpha} = (\alpha_1,\cdots,\alpha_N)$ that would, to provide \textit{reliability}, minimize the quantity $\sum_{i = 1}^N \sum_{j=1}^N \alpha_i \alpha_j \mathrm{cov}(\mathcal{R}_i, \mathcal{R}_j)$, which determines the \textit{variability} of the quality of the reconstructed scene. 
The constraints of~\eqref{cameraSelectOptProblem} are the: 
\begin{itemize}
\item \textit{quality constraint}~\eqref{eq:qualityConstraint}, which ensures that
the expected total resolution of images obtained from the cameras satisfy a given threshold $\Theta$ (to make the reconstructed scene meet quality requirements). Note that $\Theta$ controls the tradeoff between quality and reliability, 
with higher $\Theta$ values 
emphasizing selection of cameras that make the reconstructed scene to expectedly have a good quality (while sacrificing reliability) and lower $\Theta$ values reversing the trend; 
\item \textit{resource constraint}~\eqref{eq:resourceConstraint}, which ensures that the total number of selected cameras do not exceed $\Psi$ (to meet factors like bandwidth and energy
limitations); and  
\item \textit{selection parameter constraint}~\eqref{eq:probabilityFeasibility}, 
which ensures camera selection parameter definition.
\end{itemize}

\vspace{-0.1in}
\subsection{Genetic Algorithmic Solution}
\label{sec:solution}

To solve the optimization problem in~\eqref{cameraSelectOptProblem}, we employ a genetic algorithm (GA) that efficiently explores the large solution space to find optimal camera configurations, as described below: 


{\em Initial population:} The initial population is generated by randomly selecting camera configurations. Each individual represents a potential solution, encoded as a binary vector indicating camera inclusion. 

{\em Fitness function:} The fitness function evaluates each camera configuration by balancing two key factors: achieving a minimum quality threshold ($\Theta$) and minimizing the risk of correlated failures. The total quality of the selected cameras is computed as the weighted sum of their resolutions, where each camera’s importance is captured by its respective weight ($\alpha$). If the total quality falls below the threshold $\Theta$, a large penalty is assigned, ensuring only configurations meeting the minimum quality requirement are viable. If the quality exceeds $\Theta$, the risk of correlated failures is calculated based on the covariance of image resolutions between selected cameras, adjusted by their respective weights ($\alpha$). This risk minimization encourages robust configurations that reduce the likelihood of simultaneous failures. The above described fitness function is defined as:

\footnotesize
\[
\mathcal{F} =
\begin{cases}
\infty, &
\text{if } \displaystyle \sum_{i=1}^{N} \alpha_i > \Psi, \\[0.5em]
\Theta - \displaystyle\sum_{i=1}^{N} \alpha_i \,\mathbb{E}[\mathcal{R}_i], &
\text{if } \displaystyle \sum_{i=1}^{N} \alpha_i \,\mathbb{E}[\mathcal{R}_i] < \Theta, \\[0.5em]
\displaystyle\sum_{i=1}^{N} \sum_{j=i}^{N} \alpha_i \alpha_j R_i R_j \rho_{ij} \sigma_i \sigma_j, &
\text{otherwise}.
\end{cases}
\]
\normalsize


{\em GA parameters:} We define key parameters for the GA, including:
\begin{itemize}
    \item \textbf{Population size ($P_s$):} Number of individuals in the population.
    \item \textbf{Maximum generations ($G_m$):} The maximum number of iterations or generations the algorithm will run.
    \item \textbf{Crossover rate ($C_r$):} Probability of combining two parents to generate offspring.
    \item \textbf{Mutation rate ($M_r$):} Probability of altering an individual's configuration to maintain genetic diversity.
    \item \textbf{Elitism count ($E_c$):} The number of top-performing individuals retained in each generation.
    \item \textbf{Updated population ($P_{gen+1}$):} The new population for the next generation, formed by the union of elite individuals and offspring: 
{\footnotesize
\[
P_{gen+1} \gets E_c \cup \text{offspring}
\]
}

    where $E_c$ represents the elite individuals and \text{offspring} represents the children generated through crossover and mutation.
\end{itemize}

\section{Performance Evaluation}
\label{sec:finalevaluation}
Here, we discuss the performance evaluation of our proposed portfolio theoretic edge resource management technique.

\subsection{Simulation Environment}


For the evaluation, we create a detailed simulation with a realistic JIT edge environment and video dataset that is customized for multi-view 3D reconstruction.  
Most of the simulation parameters are chosen based on the dataset and typical camera specifications and operational constraints within a JIT edge environment to ensure a realistic and practical evaluation. For one of the datasets consisting of 7 simultaneous video streams, we set the number of cameras ($N$) for reconstruction to 7 with 3 cameras having a high resolution (1080p, i.e., $1920 \times 1080$ pixels) and the remaining 4 with a lower resolution (720p, i.e., $1280 \times 720$ pixels).  
For another dataset with 5 simultaneous video streams, $N$ is set to 5 with 2 cameras having high resolution.


In order to simulate the disruptions, each camera's failure probability ($p_{c_i}$) is considered to be a random variable that follows the Beta distribution. The Beta distribution is ideal for modeling probabilities due to its bounded range between 0 and 1. For the Beta distribution, parameter $x$ values are set to $[6, 6, 6, 2, 2.5, 3.5, 5]$ and parameter $y$ values are set to $[3, 3, 3, 3, 3.5, 2.5, 2]$. 
Additionally, a $7 \times 7$ correlation matrix $p_{ij}$ is used to model the correlation of camera failures, with values ranging from 0.1 to 1, capturing how disruptions in cameras statistically depend on each other. This combination of Beta-distributed probabilities and the correlation matrix ensures that the simulation realistically reflects disruption patterns and system performance under varying constraints.

\begingroup
\scriptsize
\begin{table*}[htb]
\centering
\begin{tabular}{|c|c|c|c|c|c|c|c|}
\hline
\multirow{2}{*}{Camera} & \multicolumn{2}{c|}{6 out of 7 ($\Psi^\prime=6$)} & \multicolumn{2}{c|}{5 out of 7 ($\Psi^\prime=5$)} & \multicolumn{2}{c|}{4 out of 7 ($\Psi^\prime=4$)} \\ \cline{2-7}
 & Genetic Algorithm & Optimal & Genetic Algorithm & Optimal & Genetic Algorithm & Optimal \\ \hline
1 & \xmark & \xmark & \xmark & \xmark & \cmark & \xmark \\ \hline
2 & \cmark & \cmark & \xmark & \xmark & \cmark & \xmark \\ \hline
3 & \cmark & \cmark & \cmark & \cmark & \xmark & \xmark \\ \hline
4 & \cmark & \cmark & \cmark & \cmark & \xmark & \cmark \\ \hline
5 & \cmark & \cmark & \cmark & \cmark & \cmark & \cmark \\ \hline
6 & \cmark & \cmark & \cmark & \cmark & \xmark & \cmark \\ \hline
7 & \cmark & \cmark & \cmark & \cmark & \cmark & \cmark \\ \hline
\end{tabular}
\caption{Camera Selection Comparison (\cmark\ => Selected, \xmark\ => Not selected)}
\label{table:camera_selection}
\vspace{-0.2in}
\end{table*}
\endgroup


\subsection{Datasets used}
We use popular publicly available multi-view 3D reconstruction dataset, viz., Dance1 and Odzemok \cite{Mustafa_2015_ICCV}, for performance analysis of dynamic scenes. Both are indoor datasets with cluttered backgrounds with a moving person and static background. Both datasets have 7 static cameras with resolutions as explained before. 
We also use our own multi-view dataset \cite{10.1145/3583740.3630267} \cite{10817994} that
represent different degrees of indoor dynamic scenes, e.g., Walk and Handshake. The walk dataset is moderately dynamic, where a person walks across the scene from one end to another. On the other hand, the Handshake dataset represents a high dynamism, where two people walk from two different ends of the scene and shake hands in the middle. These datasets demonstrate the proposed method's utility in providing reliable reconstruction where the 3D scene is highly dynamic. 
\subsection{Performance metrics}


For the evaluation, we measure the performance of the proposed portfolio theoretic approach by estimating how reliably it can provide useful reconstruction. Consequently, we measure the reliability of 3D reconstruction (i.e., with and without our proposed portfolio theoretic approach) by ascertaining how consistently our approach generate reconstruction with a certain minimum quality (i.e., Threshold $\Theta$). Such quality is measured by the number of vertices in the reconstructed 3D point cloud output. A higher number of vertices count signifies higher quality reconstruction, capturing greater scene details and enhancing the overall fidelity of the reconstruction.
The consistency of generating useful reconstruction is measured in terms of mean and standard deviation (S.D.) of number of vertices in the reconstructed 3D scene over different disruption structures.

For simplicity, the quality threshold ($\Theta$) is set at half of the highest resolution $(1920 \times 1080) / 2 = 1,036,800$, a common practice in computer vision for image optimization across various sets of resolutions\cite{computers8020030}.
Such threshold approximation provides a balanced estimation that ensures high-quality reconstruction while accommodating a mix of camera resolutions.
For all the following results, we use the upper bound of resource constraint ($\Psi$), i.e., $\Psi^\prime$ (from Eq.~\eqref{eq:resourceConstraint}) as a system constant.
Further, for all experiments, the mean and S.D. of reconstruction quality computed over 20 runs of Beta distributed camera failure probabilities.

\subsection{Baseline Strategies}
In this paper, we define two baseline strategies to evaluate two separate aspects of our solution. 


{\em Optimal (Brute force) vs. GA based solution:}
In order to ascertain the accuracy of GA based camera selection, we choose a baseline strategy of optimal camera selection which is generated through exhaustive search of all possible camera selection combination and choosing the one with highest combined resolution. Of course, performing such exhaustive search with larger number of cameras to find the optimal camera selection is computationally challenging and thus impractical. However, the purpose of such baseline comparison is to observe how camera selection with the proposed GA based solution performs against such optimal solution.



{\em Traditional vs. Portfolio-theoretic camera selection:}
In another set of comparison, we use a traditional method of edge resource management (specifically camera selection) where cameras with highest resolution are always selected in comparison to portfolio theoretic approach that chooses cameras based on the solution of Eq.~\eqref{eq:PortfolioThPrelimObj}.

\subsection{Optimality of genetic algorithmic solution}
In Table \ref{table:camera_selection}, we present the comparison between 
the camera selection strategy obtained by solving Eq.~\eqref{eq:PortfolioThPrelimObj} using our GA-based approach and the one obtained by solving it using a brute force method.
While the brute force method evaluates all possible combinations exhaustively, 
the GA employs evolutionary principles, such as selection, crossover, and mutation, to efficiently explore the solution space. In Table \ref{table:camera_selection}, we observe that for different camera selection scenarios (i.e., different $\Psi$), 
the solutions provided by the two are predominantly the same.
Such consistency validates GA's effectiveness in solving the optimization problem. One also has to note that although the GA seeks to find the near-optimal solution, it is after all an approximation strategy which can result in a solution that is slightly different from the optimal solution. However, what the proposed GA sacrifices in accuracy, it more than makes it up by employing an iterative approach that can handle larger problem sizes (i.e., number of cameras in our case) more feasibly, providing near-optimal solutions with significantly reduced computational requirements.  Such consistency also offers a practical justification for using the GA in real-world applications, as it reconciles the exhaustive search with the iterative approach, ensuring reliable and relevant outcomes. 

\begin{figure}[htbp]
  \centering
  \subcaptionbox{(a) Dance1 \& Odzemok\label{fig:meandance1-tight}}%
    [0.49\linewidth]{%
      \includegraphics[width=\linewidth,trim=10 6 10 6,clip]{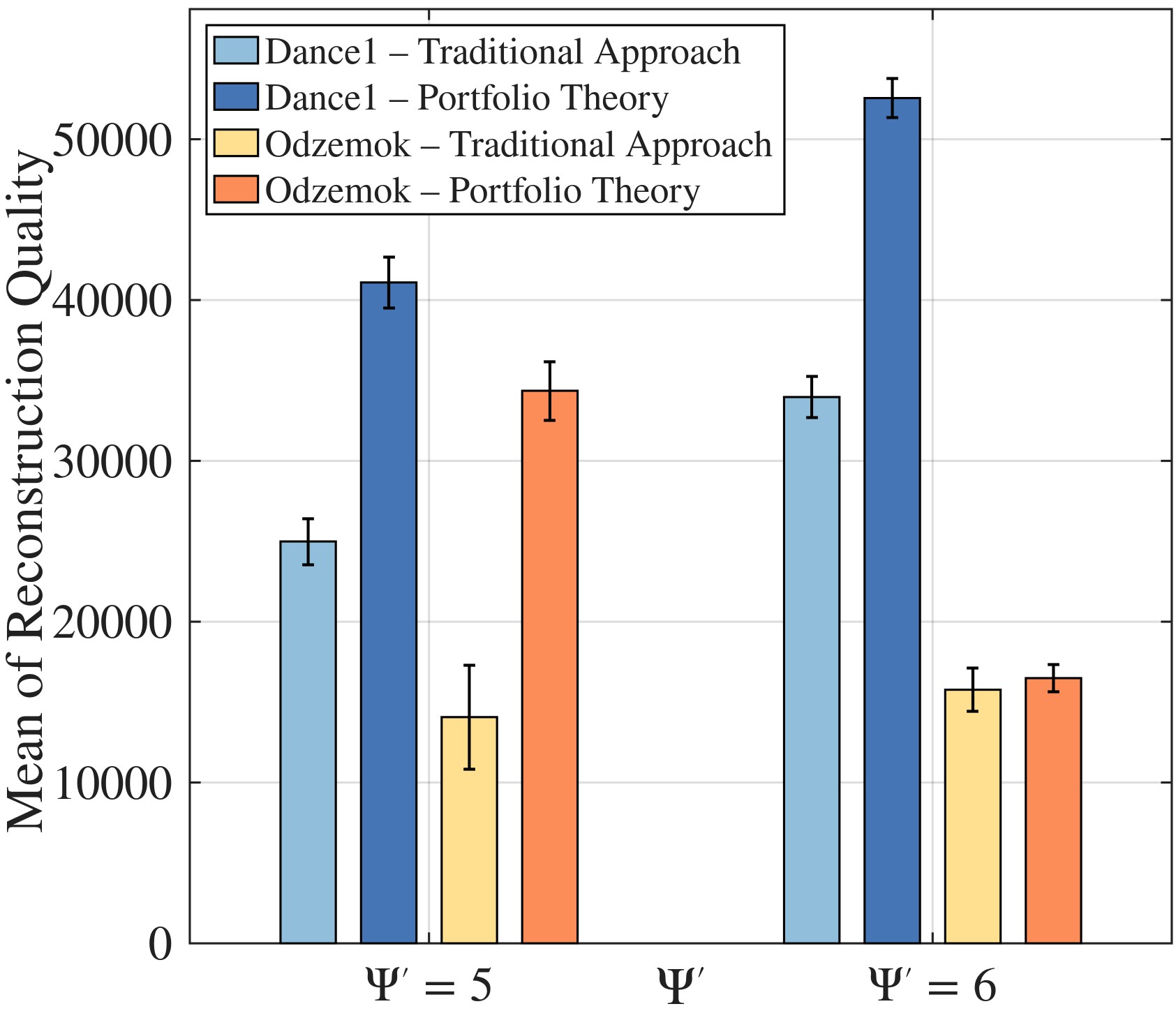}%
    }%
  \hfill%
  \subcaptionbox{(b) Handshake \& Walk\label{fig:meanodzemok-tight}}%
    [0.49\linewidth]{%
      \includegraphics[width=\linewidth,trim=10 6 10 6,clip]{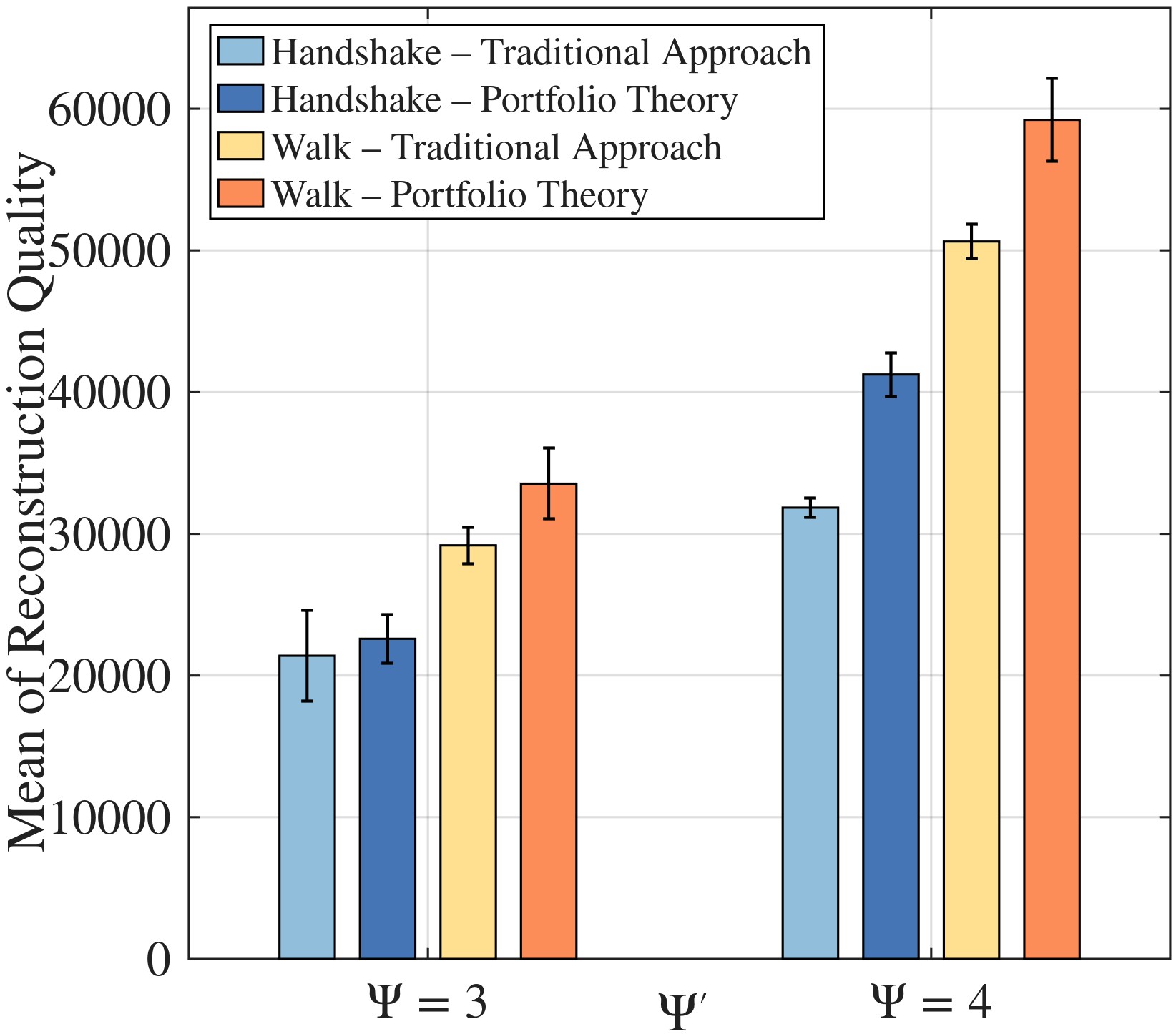}%
    }%
  \vspace{2pt}
  \caption{Mean of Reconstruction Quality for Traditional vs.\ Portfolio.}
  \label{fig:mean_recon_quality_tight}
  \vspace{-4pt}
\end{figure}

\subsection{Reliability Comparison Against Traditional Approach}
Here we present comparative results between two approaches —Portfolio Theoretic camera selection and Traditional Approach towards camera selection—in terms of their performance in ensuring reliable reconstruction under disruptions. As explained earlier, such reliability is measured in terms of the approaches' ability to ensure consistently high quality reconstruction over threshold ($\Theta$). Further, in the following results, mean and standard deviation (S.D.) of reconstruction quality (in terms of number of vertices) are measured over multiple images sets (each image set is a set of images from different cameras belonging to the same timestamp) belonging to different datasets. Choosing multiple image sets from the same dataset examines how the proposed portfolio theoretic camera selection strategy preforms (in terms of generating reliable reconstruction) over a longer peroid
as each dataset contains synchronized images from multiple cameras over a long term period in terms of streams of images. While choice of multiple datasets investigates the proposed method's versatility in dealing with 3D scenes that have a varied degree of dynamism as explained earlier.   


\subsubsection{Mean of reconstruction quality for different disruption structures}
Figure \ref{fig:mean_recon_quality_tight} displays the mean reconstruction quality (in terms of number of vertices) over multiple image sets across datasets as a function of the upper bound on the number of cameras to be used for reconstruction ($\Psi^\prime$). Error bars represent the variance across 20 different runs, providing insight into the consistency of each method.

In Figure \ref{fig:meandance1-tight} (Dance1 \& Odzemok), \emph{Dance1} shows a clear gain: at $\Psi^\prime\!=\!5$ Portfolio Theory is $\sim$40{,}000 vs.\ $\sim$25{,}000 for the Traditional Approach; at $\Psi^\prime\!=\!6$ Portfolio Theory reaches $\sim$52{,}000 vs.\ $\sim$33{,}000. For \emph{Odzemok}, at $\Psi^\prime\!=\!5$ Portfolio Theory is $\sim$34{,}000 vs.\ $\sim$15{,}000 for Traditional, whereas at $\Psi^\prime\!=\!6$ both methods are much lower (Traditional $\sim$16{,}000, Portfolio $\sim$17{,}000), with only a small gap. Across both datasets, Portfolio Theory generally exhibits tighter error bars.

In Figure \ref{fig:meanodzemok-tight} (Handshake \& Walk), \emph{Handshake} shows a modest advantage at $\Psi^\prime\!=\!3$ (Portfolio $\sim$24{,}000 vs.\ Traditional $\sim$22{,}000) that widens at $\Psi^\prime\!=\!4$ (Portfolio $\sim$41{,}000 vs.\ $\sim$32{,}000). \emph{Walk} shows a consistent and larger gain: at $\Psi^\prime\!=\!3$ Portfolio is $\sim$34{,}000 vs.\ $\sim$29{,}000, and at $\Psi^\prime\!=\!4$ Portfolio is nearly $\sim$60{,}000 vs.\ $\sim$51{,}000 for Traditional, again with narrower error bars for Portfolio Theory.

\subsubsection{S.D. of reconstruction quality for different disruption structures}
\begin{figure}[htbp]
  \centering
  \subcaptionbox{(a) Dance1 \& Odzemok\label{fig:sddance1}}%
    [0.49\linewidth]{%
      \includegraphics[width=\linewidth,trim=10 6 10 6,clip]{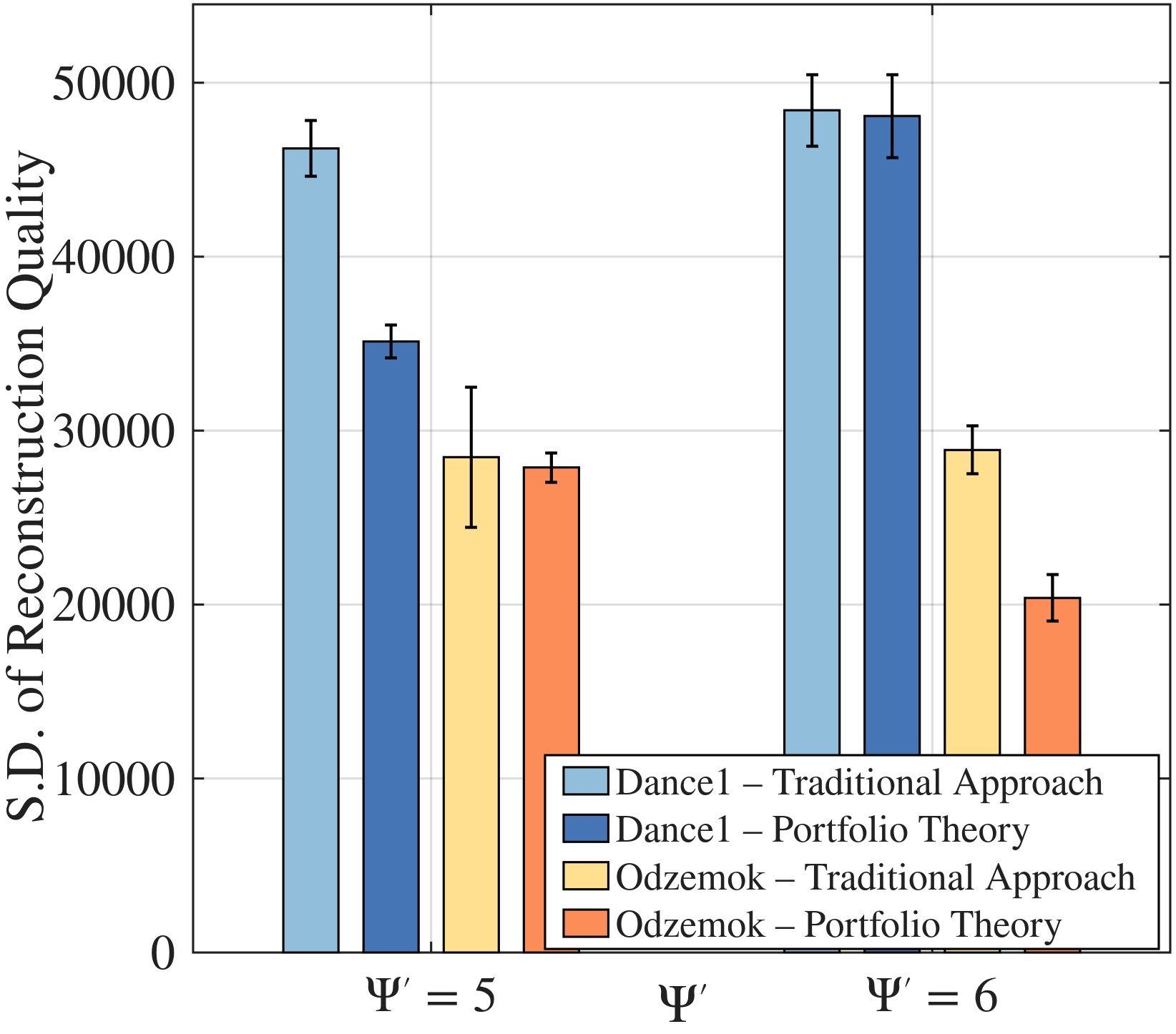}%
    }%
  \hfill%
  \subcaptionbox{(b) Handshake \& Walk\label{fig:sdhandshakewalk}}%
    [0.49\linewidth]{%
      \includegraphics[width=\linewidth,trim=10 6 10 6,clip]{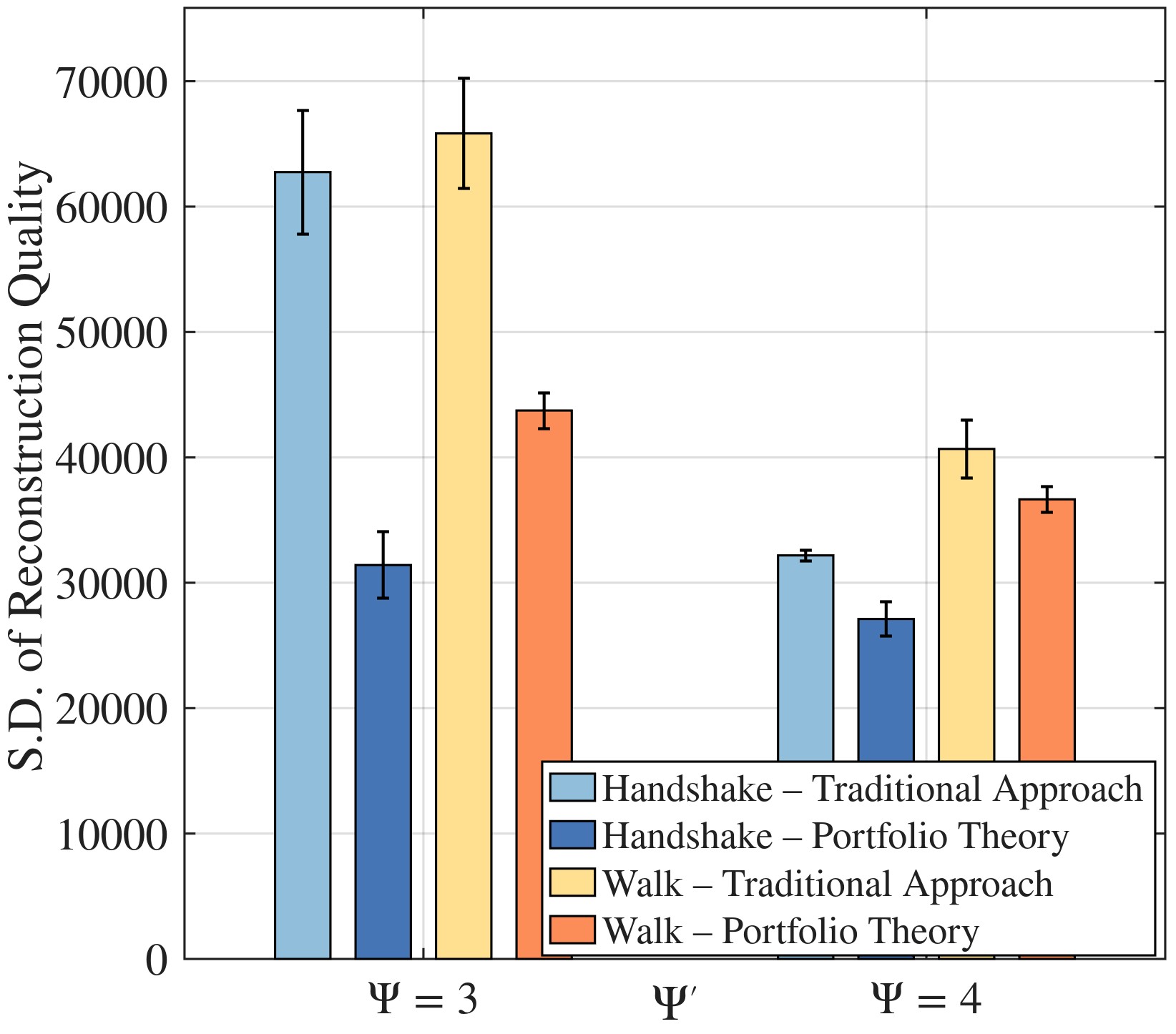}%
    }%
  \vspace{2pt}
  \caption{Reconstruction S.D. of Reliability for Traditional vs.\ Portfolio.}
  \label{fig:sd_reliability_tight}
  \vspace{-4pt}
\end{figure}

This section presents a quantitative comparison between Portfolio Theory and the Traditional Approach in terms of reconstruction reliability, measured by the standard deviation (S.D.) of reconstruction quality across different image sets in four datasets. Figure~\ref{fig:sd_reliability_tight} shows the S.D. values as a function of $\Psi^\prime$, with error bars representing variability over 20 different runs.

In Figure~\ref{fig:sddance1} (Dance1 \& Odzemok), \emph{Dance1} shows a clear reduction in variability: at $\Psi^\prime\!=\!5$ the Traditional Approach records about $47{,}000$, compared to $\sim\!35{,}000$ for Portfolio Theory; at $\Psi^\prime\!=\!6$ the gap widens, with Traditional near $49{,}000$ and Portfolio below $30{,}000$. For \emph{Odzemok}, at $\Psi^\prime\!=\!5$ the S.D.\ is $\sim\!28{,}000$ for Traditional vs.\ $\sim\!28{,}000$ for Portfolio Theory (nearly equal), while at $\Psi^\prime\!=\!6$ Traditional stays near $29{,}000$ and Portfolio drops to about $20{,}000$. Across both datasets, Portfolio Theory tends to produce smaller error bars, indicating greater consistency.

In Figure~\ref{fig:sdhandshakewalk} (Handshake \& Walk), \emph{Handshake} shows a large difference at $\Psi^\prime\!=\!3$, with Traditional at $\sim\!63{,}000$ and Portfolio near $31{,}000$; at $\Psi^\prime\!=\!4$, Traditional is $\sim\!32{,}000$ and Portfolio $\sim\!28{,}000$. For \emph{Walk}, at $\Psi^\prime\!=\!3$ the Traditional Approach records $\sim\!66{,}000$ vs.\ $\sim\!42{,}000$ for Portfolio Theory; at $\Psi^\prime\!=\!4$ both are closer (Traditional $\sim\!39{,}000$, Portfolio $\sim\!37{,}000$) but Portfolio still shows slightly less variability.

\begin{figure}[htbp]
  \centering
  \subcaptionbox{(a) Dance1 \& Odzemok\label{fig:rnkdance1}}%
    [0.49\linewidth]{%
      \includegraphics[width=\linewidth,trim=10 6 10 6,clip]{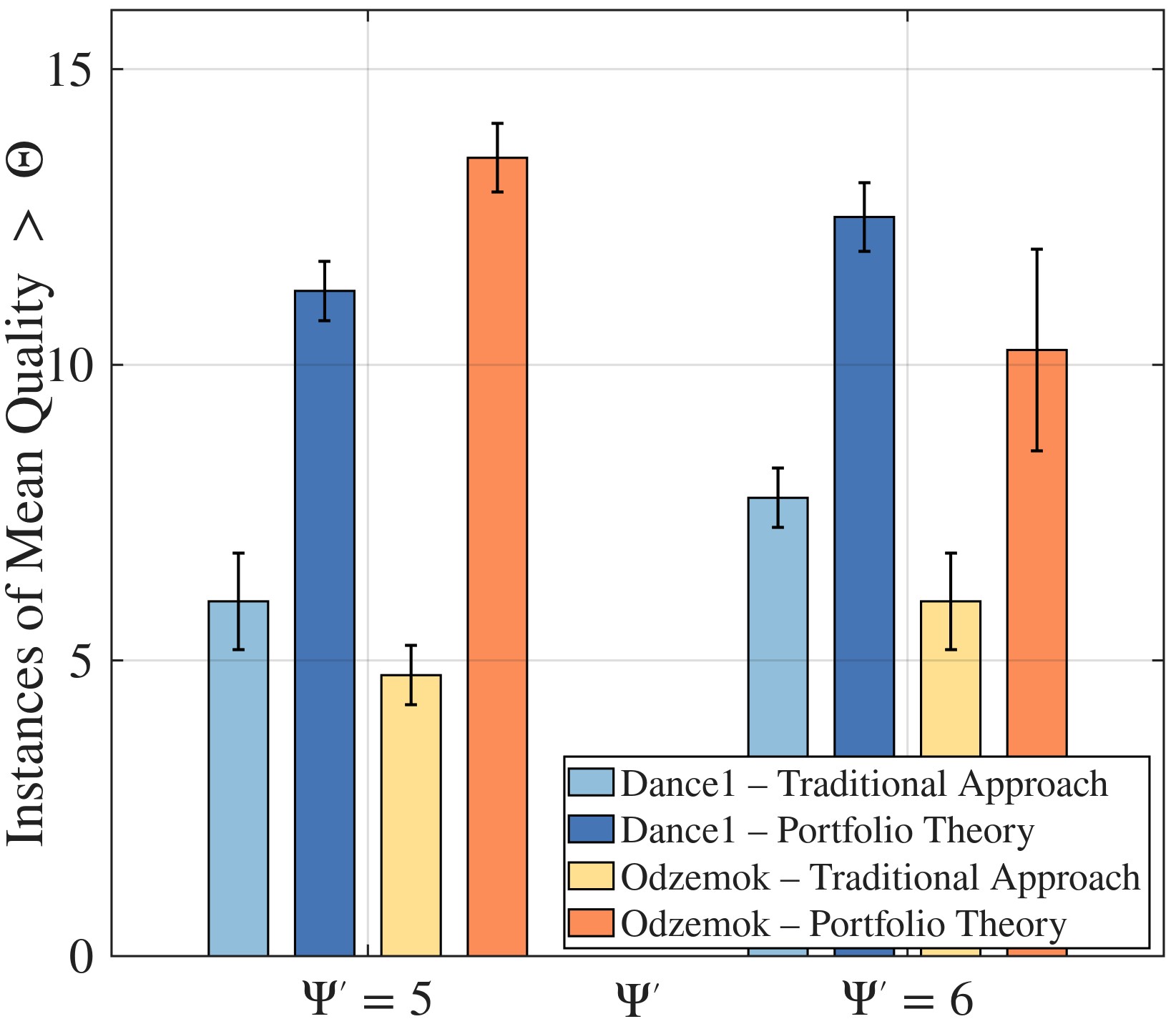}%
    }%
  \hfill%
  \subcaptionbox{(b) Handshake \& Walk\label{fig:rnkhandshakewalk}}%
    [0.49\linewidth]{%
      \includegraphics[width=\linewidth,trim=10 6 10 6,clip]{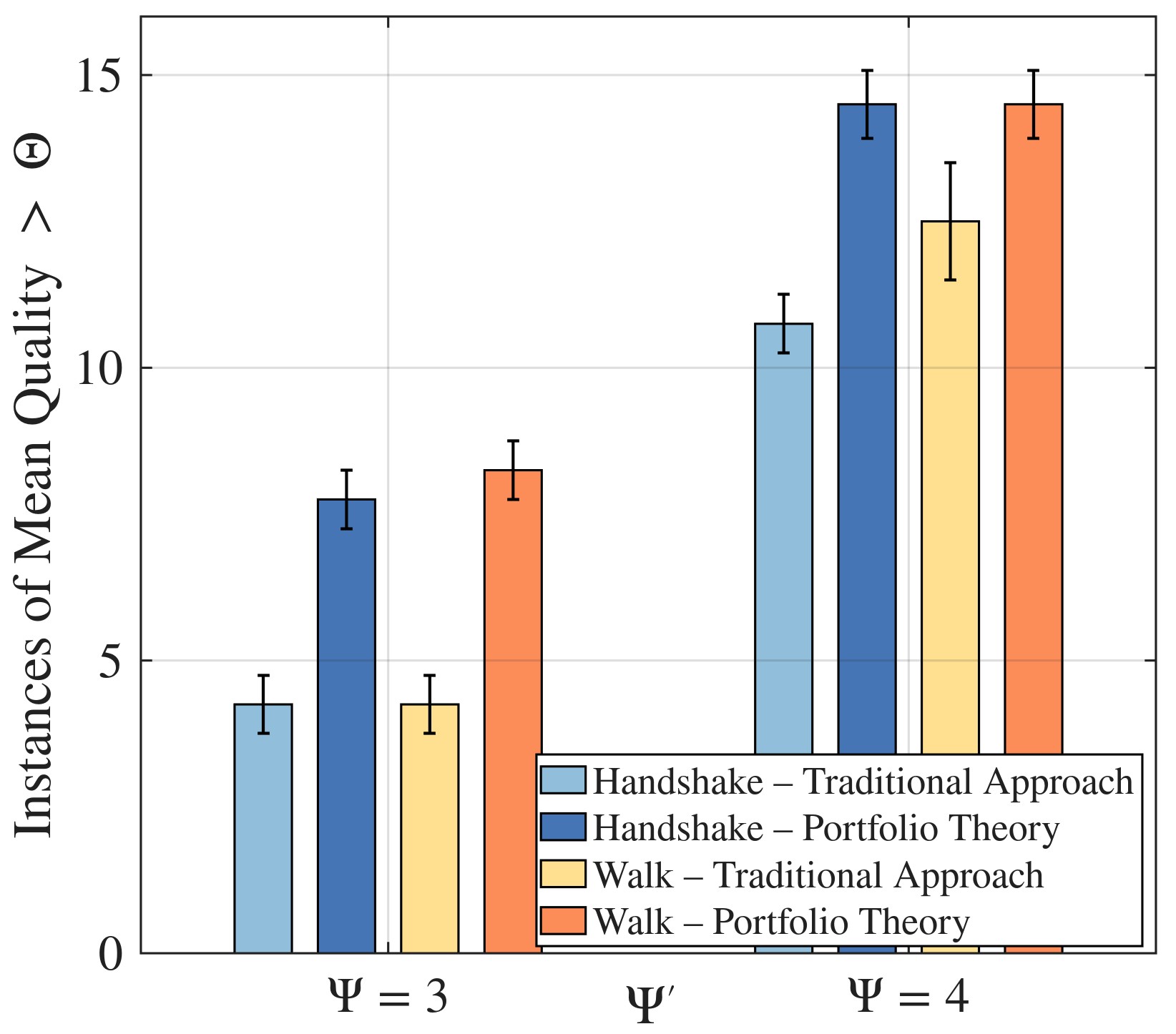}%
    }%
  \vspace{2pt}
  \caption{Reconstruction Quality Over Threshold for Traditional vs.\ Portfolio.}
  \label{fig:quality_reliability_tight}
  \vspace{-4pt}
\end{figure}

\subsubsection{Mean reconstruction quality above $\Theta$ for different disruption structures}
In this section, we evaluate the number of instances where the mean reconstruction quality surpasses a threshold $\Theta$ across four datasets. Figure~\ref{fig:quality_reliability_tight} shows these instances as a function of $\Psi^\prime$, with error bars indicating variability over 20 runs and highlighting consistency.

In Figure~\ref{fig:rnkdance1} (Dance1 \& Odzemok), \emph{Dance1} at $\Psi^\prime = 5$ records about 6 instances for the Traditional Approach and 12 for Portfolio Theory; at $\Psi^\prime = 6$ these rise to roughly 8 and 13, respectively. For \emph{Odzemok}, at $\Psi^\prime = 5$ the Traditional Approach has around 5 instances versus 14 for Portfolio Theory; at $\Psi^\prime = 6$ the values are about 6 and 10. In both datasets, Portfolio Theory generally shows smaller error bars, indicating greater consistency.

In Figure~\ref{fig:rnkhandshakewalk} (Handshake \& Walk), \emph{Handshake} at $\Psi^\prime = 3$ achieves about 4 instances for the Traditional Approach compared to 7 for Portfolio Theory; at $\Psi^\prime = 4$ the counts are roughly 11 and 14. For \emph{Walk}, at $\Psi^\prime = 3$ the Traditional Approach records about 5 instances versus 8 for Portfolio Theory; at $\Psi^\prime = 4$ these increase to about 13 and 15, respectively. Across both datasets, Portfolio Theory maintains narrower error bars, reflecting its stronger reliability.

\begin{figure*}[t]
    \centering

    \hspace*{\fill} 
    \raisebox{0.05cm}{\includegraphics[width=0.14\textwidth]{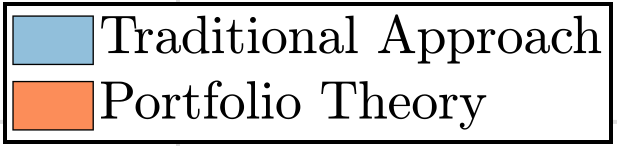}} 

    \vspace{-0.02cm} 

    \begin{subfigure}{0.24\textwidth}
        \centering
        \includegraphics[width=\linewidth, height=\linewidth]{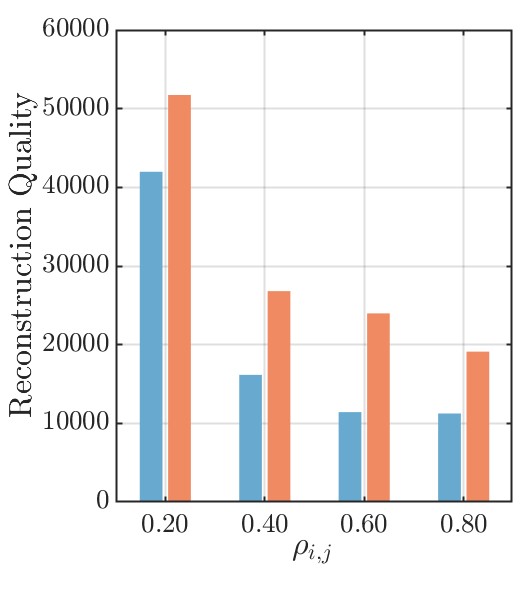}  
        \vspace{-0.2in}
        \caption{(a) Dance1} 
        \label{fig:corrmeandance1} 
    \end{subfigure}
    \hfill
    \begin{subfigure}{0.24\textwidth}
        \centering
        \includegraphics[width=\linewidth, height=\linewidth]{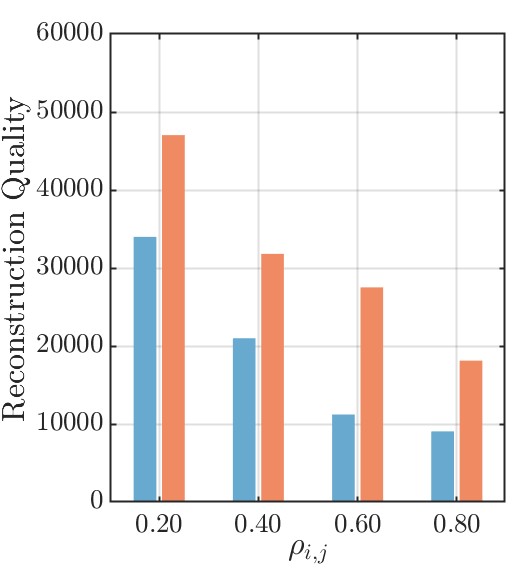}  
        \vspace{-0.2in}
        \caption{(b) Odzemok} 
        \label{fig:corrmeanodzemok} 
    \end{subfigure}
    \hfill
    \begin{subfigure}{0.24\textwidth}
        \centering
        \includegraphics[width=\linewidth, height=\linewidth]{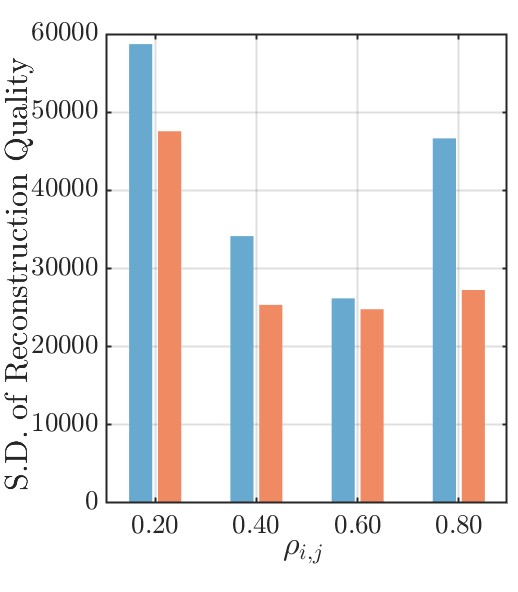}  
        \vspace{-0.2in}
        \caption{(c) Dance1} 
        \label{fig:corrsddance1} 
    \end{subfigure}
    \hfill
    \begin{subfigure}{0.24\textwidth}
        \centering
        \includegraphics[width=\linewidth, height=\linewidth]{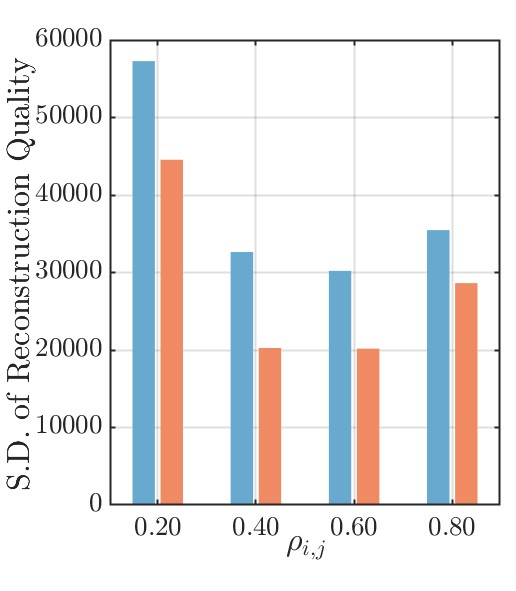}  
        \vspace{-0.2in}
        \caption{(d) Odzemok} 
        \label{fig:corrsdodzemok} 
    \end{subfigure}
    \caption{Impact of Varying $\rho_{i,j}$ on Reconstruction Quality}
    \label{fig:correlation_impacts}
    \vspace{-0.2in}
\end{figure*}

\subsection{Impact of Disruption Correlations on Performance}

Here, in Figures~\ref{fig:correlation_impacts} and~\ref{fig:correlation_side_by_side}, we study how the degree of correlations ($\rho_{i,j}$) among disruptions experienced by cameras impact 3D reconstruction performance 
using the \textit{Dance1} and \textit{Odzemok} datasets. Reconstruction reliability, measured through the mean vertices metric, demonstrates that a portfolio theory-based approach significantly outperforms traditional methods. By optimizing the trade-off between resolution and dependability, the portfolio theory approach exhibits consistent reliability across diverse correlation levels, underscoring its potential to advance 3D image reconstruction techniques for real-time, high-resolution applications.

\subsubsection{Reconstruction Reliability}

The analysis of mean reconstruction reliability across the Dance1 and Odzemok datasets highlights the consistent superiority of the portfolio theory compared to the traditional approach, particularly as the correlation among higher-resolution cameras (\(\rho_{i,j}\)) increases. For the Dance1 dataset (Figure~\ref{fig:corrmeandance1}), the traditional approach exhibits a sharp decline in reconstruction quality as \(\rho_{i,j}\) increases from 0.2 to 0.8. At \(\rho_{i,j} = 0.2\), the traditional approach achieves a mean reconstruction quality of approximately 30,000, while the portfolio theory significantly outperforms it with a quality exceeding 50,000. As \(\rho_{i,j}\) increases, the traditional approach's quality plummets to below 15,000 at \(\rho_{i,j} = 0.8\), whereas the portfolio theory demonstrates remarkable stability, retaining a quality of nearly 25,000 even under high correlation conditions.

A similar pattern is observed in the Odzemok dataset (Figure~\ref{fig:corrmeanodzemok}). The traditional approach begins with a quality of around 25,000 at \(\rho_{i,j} = 0.2\) but declines sharply, reaching below 10,000 at \(\rho_{i,j} = 0.8\). In contrast, the portfolio theory starts with a quality exceeding 40,000 at \(\rho_{i,j} = 0.2\) and maintains a significant advantage across all correlation levels, achieving approximately 20,000 at \(\rho_{i,j} = 0.8\). These results underscore the robustness of the portfolio theory in managing high-resolution camera data with varying correlation levels. Its ability to maintain superior reconstruction quality and reliability, even under challenging conditions, highlights its potential for real-time 3D reconstruction tasks where consistency and performance are critical.

\subsubsection{Reconstruction S.D. of Reliability}

Analyzing the standard deviation (SD) of reconstruction quality across the Dance1 and Odzemok datasets provides valuable insights into each method’s robustness. A higher SD indicates greater variability and, consequently, reduced reliability. For the Dance1 dataset (Figure~\ref{fig:corrsddance1}), as the correlation \(\rho\) increases from 0.2 to 0.8, the traditional approach exhibits a significant rise in SD, signaling diminishing robustness. In contrast, the portfolio theory approach maintains a consistently lower SD, highlighting its ability to ensure stable and reliable reconstruction across varying correlation levels.

Based on the Odzemok dataset (Figure~\ref{fig:corrsddance1}), the findings confirm the Dance1 patterns. The traditional approach shows that as correlation increases, robustness declines, indicating an increase in variability in dependability with a more significant standard deviation. In contrast, the portfolio theory demonstrates a significant standard deviation consistency at different correlation values. The technique's inherent robustness is highlighted by the slight increase in standard deviation, even at high correlation levels. 

\begin{figure}[t]
    \centering

    \hspace*{\fill} 
    \raisebox{0.24cm}{\includegraphics[width=0.15\textwidth]{Figs/Legend.jpg}} 
    \vspace{-0.2cm} 

    \begin{subfigure}{0.24\textwidth}
        \centering
        \includegraphics[width=\linewidth]{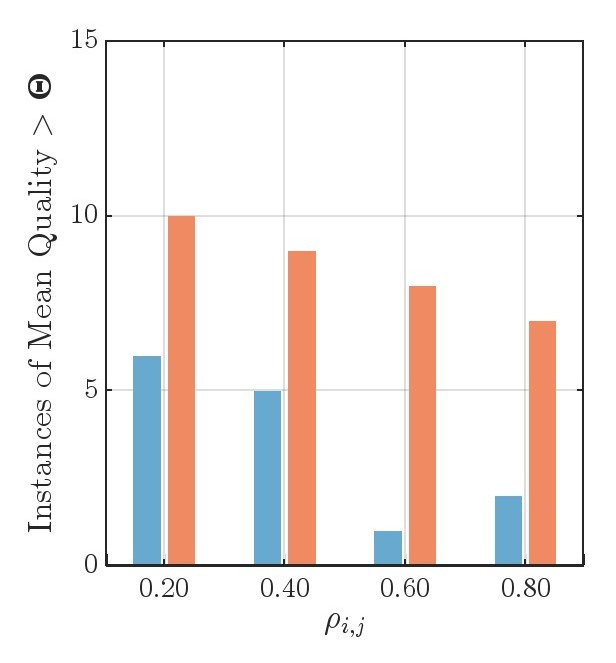}
        \vspace{-0.2in}
        \caption{(a) Dance1}
        \label{fig:corrrnkdance1}
    \end{subfigure}
    \hfill
    \begin{subfigure}{0.24\textwidth}
        \centering
        \includegraphics[width=\linewidth]{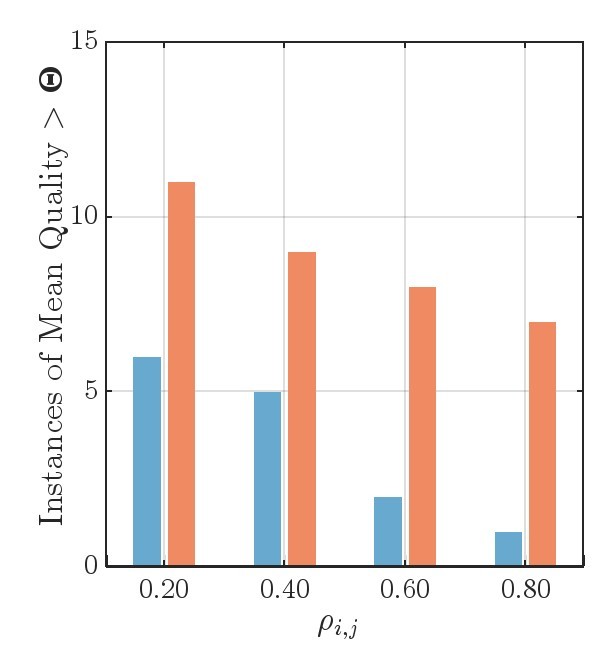}
        \vspace{-0.2in}
        \caption{(b) Odzemok}
        \label{fig:corrrnkodzemok}
    \end{subfigure}

    \caption{Impact of Varying $\rho_{i,j}$ on Reconstruction Reliability.}
    \label{fig:correlation_side_by_side}
    \vspace{-0.3in}
\end{figure}

\subsubsection{Reconstruction Quality Over Threshold}

The analysis of the Dance1 dataset (Figure~\ref{fig:corrrnkdance1}) demonstrates the portfolio theory's consistent superiority over the traditional approach across varying correlation levels (\(\rho\)). At lower correlation levels (\(\rho = 0.2\)), the portfolio theory achieves substantially higher reconstruction quality, indicating its exceptional performance when camera outputs exhibit minimal similarity. As \(\rho\) increases, both methods experience a decline in quality, yet the portfolio theory consistently maintains an edge over the traditional approach, confirming its robustness in ensuring higher-quality reconstructions.

Similarly, in the Odzemok dataset (Figure~\ref{fig:corrrnkodzemok}), portfolio theory consistently outperforms the traditional approach, particularly at lower correlation levels (\(\rho = 0.2\)). Even at higher correlations (\(\rho = 0.6\) and \(\rho = 0.8\)), portfolio theory sustains a higher quality threshold, underscoring its capability to deliver superior reconstruction quality under challenging conditions. These results make portfolio theory the preferred choice for high-resolution, real-time 3D reconstruction applications requiring resilience to fluctuating camera correlations.

Our analysis reveals a clear pattern where the portfolio theory consistently outperforms the traditional approach to provide superior reconstruction quality and reliability in the presence of varying degrees of correlated disruptions. This consistently superior performance highlights the potential of the portfolio theory to deliver high-quality outputs for real-time applications, making it the preferred choice for systems that demand reliable quality in dynamic environments.

\section{Conclusions}
\label{sec:conclusions}

In this work, we introduced a novel portfolio-theoretic approach to optimize 3D reconstruction scenes, effectively balancing trade-offs between quality and reliability. Experimental results using real datasets demonstrated significant performance advantages over traditional methods, and the diverse datasets highlighted the scalability of our approach. We also proposed a genetic algorithm for efficiently finding camera selection strategies from our portfolio-theoretic perspective. Looking ahead, we will develop a comprehensive system-wide model that will integrate advanced technologies such as deep learning, system monitoring, and constrained optimization, aiming to understand and model spatiotemporally correlated disruptions while formulating a management strategy that will adaptively respond to these disruptions among cameras. Additionally, we will analytically quantify the trade-offs between quality, latency, and reliability under disruptions, ultimately striving to measure these trade-offs using advanced deep neural networks.

{\bibliographystyle{IEEEtran}}
\bibliography{paper}
\end{document}